# HENet: A Highly Efficient Convolutional Neural Networks Optimized for Accuracy, Speed and Storage

Qiuyu Zhu
Shanghai University
zhuqiuyu@staff.shu.edu.cn

Ruixin Zhang
Shanghai University
chriszhang96@shu.edu.cn

## Abstract

In order to enhance the real-time performance of convolutional neural networks (CNN), more and more researchers are focusing on improving the efficiency of CNN. Based on the analysis of some CNN architectures, such as ResNet, DenseNet, ShuffleNet and so on, we combined their advantages and proposed a very efficient model called Highly Efficient Networks(HENet). The new architecture uses an unusual way to combine group convolution and channel shuffle which was mentioned in ShuffleNet. Inspired by ResNet and DenseNet, we also propose a new way to use element-wise addition and concatenation connection with each block. In order to make greater use of feature maps, pooling operations are removed from HENet. The experiments show that our model's efficiency is more than 1 times higher than ShuffleNet on many open source datasets, such as CIFAR-10/100 and SVHN. Code is available at https://github.com/anlongstory/HENet

## 1. Introduction

The achievements of LeNet [1] and AlexNet [2] laid the standard structural pattern for convolution neural networks, a series of convolution layers followed by some fully connected layers. As time went by, researchers began to explore the effects of depth and width of the neural network. Based on the 2014 ImageNet [3] two classic network VGG16[4] and GoogLeNet [5] are produced. Then, modularization and small convolution filters are come into view. Start from the Inception structure proposed in GoogLeNet, the neural network developed towards the direction of modularization, and the traditional structure was gradually get rid of. Modularity can make networks more flexible, and the decomposition from large convolution kernel to small kernel can reduce the number of the network parameters and increase the depth of the network, so as to increase the network's nonlinearity. Subsequently, ResNet [6] proposed in 2015, as the first network, surpass the human performance in the ImageNet Large Scale Visual Recognition Challenge(ILSVRC), whose cross-layer connection gives us a new way to design more effective network. Although the primary trend of CNN is deeper and larger, but real-time application is an ineluctable problem. We all know that with the increasing of network depth, parameters become larger. How to make the network parameters more effective, and to improve the efficiency of the implementation of the network are still a problem to be solved.

Depthwise separation convolution is the key to many effective networks, in which MobileNet [7] and ShuffleNet [8] are the state-of-the-art. Although the depthwise separation convolution can reduce a certain amount of computation, accelerate the network, but this method requires network unit or block multiple repeats to reduce its impact on accuracy, especially for small networks. For some embedded devices with limited computational power, such as Raspberry Pi, many times of block repeat pose a new problem: the number of layers in network has increased, which means the additional allocation of memory, and data processing. The bottleneck structure of ResNet has been accepted by many researchers, but the bottleneck structure seems less important in the highly redundant network structure [9].

To solve these problems, this paper proposes a new network structure, which mainly adopt different number of groups within group convolution layer, and the number of groups depends on the amount of input channels of each convolution layer. In each block we use element-wise addition operation and concatenation operation for cross layer connection. In order to show the effectiveness of HENet in case of small networks, we experimented with small architecture and trained on small-size input datasets. Thus it is convenient for us to obtain the experimental results quickly.

The main contributions of this paper:
1. The amount of group changes with the input channels, and the corresponding rules are designed.
2. Proposing a new skip connection way that combines the characteristics of ResNet and DenseNet [11], which make the network implementation more efficient.
3. Based on the above two points and other methods of improving network efficiency, a new network structure HENet is proposed, which has more efficient network performance.

The content of the paper is arranged as follows. Section 2 reviews the related work. Section 3 gives an introduction to HENet in detail. Section 4 shows the experimental results and network performance analysis. Section 5 summarizes the thesis content.

## 2. Related Works

In recent years, some effective networks are proposed, researchers have varies of ways to speed up the network. The network pruning method is used in [12, 13] to reduce network parameters, in which some small-weight, which is not sensitive to the result, are pruned after the network training. [14] uses a distillation method, in which a large "teacher" network is used to train a small "student" network. [7,8,9,15] are designed all from the view of network structure, some of them are related. [7,9,15] are all Google's work, which have the depthwise separation convolution in different forms. The group convolution was first proposed in AlexNet which was trained with two GPUs, because of the limited computational resources at that time. They have proved that this way had some benefit for precision. [7, 8] achieved the effect of increasing network efficiency by splitting the 3×3 convolution layer into smaller network structure, called building block. [7] increased efficiency mainly through group convolution and depthwise separation convolution, whose building block is shown in Fig 1 (b). [8] was based on the bottleneck structure of ResNet, and to solve the problem of 1×1 convolution calculation. A group operation is adopted, and the channel shuffle operation is added to address the side effect of multiple group convolutions. The units of ShuffleNet are shown in Fig 1 (c). The building block structure of this paper is presents in Fig 1 (a).

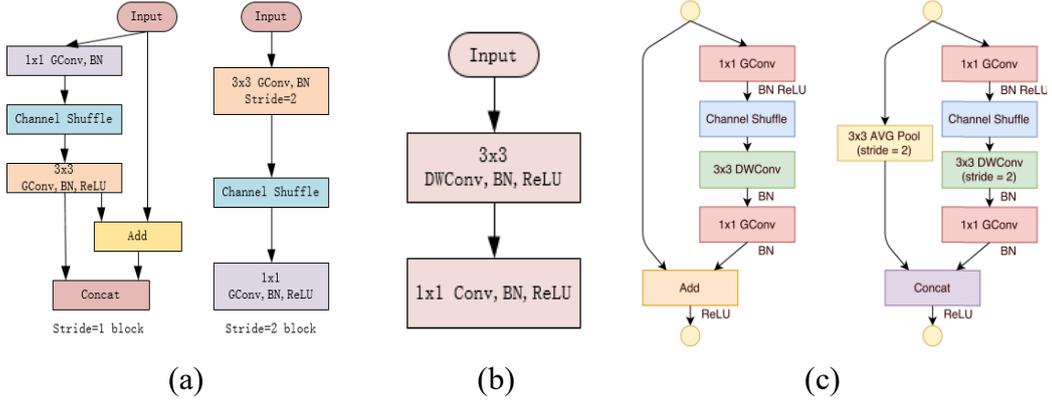

(a) (b) (c)

**Figure 1:** Comparison of different network structures. "GConv" represents group convolution, and "DWConv" stands for depthwise separation convolution. (a) HENet building blocks. (b) MobileNet block with the depthwise separation convolution. (c) ShuffleNet units which combined with group convolutions and channel shuffle. Best in color.

## 3. Model architecture

We present the building block structure of HENet in Fig 1 (a), and network parameters in Table 1. The specific structure of the network (in repeat 3 for example) is shown in Table 2.

**Table 1:** The relationship among input, output and number of group. H, W refers to the height and width of input feature map, respectively; C is the number of channels, and S stands for the stride of convolution layer.

| Stage | Input | Kernel Size | Output | Group |
|---|---|---|---|---|
| Stride 1 block | $H \times W \times 2C$ | $1 \times 1$  S=1 | $H \times W \times C$ | m |
| | | Channel shuffle | | m |
| | $H \times W \times C$ | $3 \times 3$  S=1 | $H \times W \times C$ | n |
| Stride 2 block | $H \times W \times 2C$ | $3 \times 3$  S=2 | $\frac{H}{2} \times \frac{W}{2} \times C$ | m |
| | | Channel shuffle | | m |
| | $\frac{H}{2} \times \frac{W}{2} \times C$ | $1 \times 1$  S=1 | $\frac{H}{2} \times \frac{W}{2} \times 2C$ | n |

## 3.1 Group Convolutions

AlexNet [2] trained on 2 GPUs set the beginning of the group convolution prototype. Subsequently, depthwise separation convolution is proposed, with which many of the outstanding network has been proposed [6,7,8,15] to trade off performance and efficiency. They are all inclined to adopt a unified structure unit including the same groups to build the network structure, which is well understood that unified unit is more convenient to expand. [8] solves the side effect of [6,7,15] in the process of applying multiple group convolution and addresses the problem that 1×1 convolution layer also has a large computation cost. We note that the 3 groups used in ShuffleNet is a better answer determined by many experimental comparisons. Different number of groups has different effects on performance, which require many experiments to be finalized. The use of the depthwise separation convolution reduces computation cost significantly, but for small networks in some application scenarios, reducing the number of repeats, then using the depthwise separation convolution will undoubtedly damage accuracy.

To address the above two issues, we discard the depthwise separation convolution, as shown in Fig 1 (a). The structure of the 1×1 group convolution + channel shuffle + 3×3 group convolution is used in place of the standard 3×3 convolution. The rules for setting the groups of each block layer, are given in our method.

As shown in Table 1, we combine the depthwise separation convolution into a group convolution. To balance the complexity, we designed different number of groups within the same block to make the feature maps fully blended and the features to be reused. In Table 1, $m$ and $n$ are the nearest two divisors of C, which satisfies:

$$m \times n = C \quad \text{and} \quad m > n. \tag{1}$$

For larger groups, each group will have fewer channels and computation cost. So we are going to select two of the submultiples in *C* to ensure that both of them can be as large as possible in the case of different groups, which means that *m, n* depends on the input channels per layer and once the number of input channels is determined, they are fixed. For small networks, the number of channels is rather small, larger groups need network use feature maps more effectively.

As shown in Fig 2 (b), according to our method, we set GConv1 with *m* groups, then use channel shuffle operation in *m* groups. Subsequently, GConv2 is split into *n* groups, which will contains all the above different channels of the feature maps after shuffle operation. Each GConv2 group has more channels than GConv1 group, which ensures the full use of the feature map to improve the accuracy.

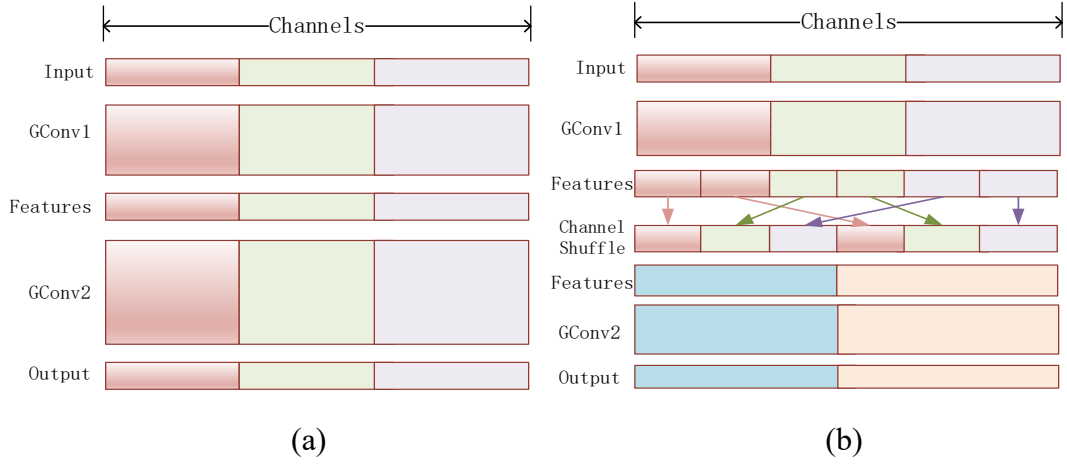

(a)                  (b)

**Figure 2:** Different group convolution situations. "GConv" stands for group convolution. (a) two stacked convolution layers have the same number of groups, channels within each input group is only related to the same color output group, and there is no information interaction between each input group. (b) two stacked convolution layers have different groups, use channel shuffle to increase the information interaction between groups. Best in color.

### 3.2 Skip Connection

Let us assume that a network contains $L$ layers, $B_0$ is input image data, and the output of layer is $B_\ell$, each block uses a non-linear transformation $H_\ell(\cdot)$, where $\ell$ means the index of the layer. $H_\ell(\cdot)$ can be a series of different operations, such as convolution, pooling, activation function, and batch normalization [10]. Traditional forward propagation network, the connection between layers is one by one, the output of $\ell\text{-}1^{th}$ layer $B_{\ell-1}$ is the input of $\ell^{th}$. We can obtain $B_\ell$ as:

$$B_\ell = H_\ell(B_{\ell-1}). \qquad (2)$$

ResNet proves that the network is no better as the layers just stack more. So, it focuses on how to increase the number of network layers, and not to cause the network performance worse, whose skip connection gives us a new idea to build neural network. Element-wise addition layer is used to combine the features from many previous layers, not just from the previous layer of the network, which undoubtedly increases the use of features and favor gradient flow. However, it may hinder the transfer of information through the network. Its structural connection is shown in Fig 3 (a). The expression of the identity transformation proposed by ResNet can be expressed as:

$$B_\ell = H_\ell(B_{\ell-1}) + B_{\ell-1}. \qquad (3)$$

DenseNet transfers all add connection, which is used in ResNet, into concatenation, so that the input of each layer is all the output of previous layers. As the network deepen, DenseNet utilizes 1×1 convolution to reduce the dimension of feature maps, and the resolution reduction is done by pooling layer. It improves the transmission of information in the network. However, because the characteristics of concatenating each layer, different layers may have similar feature maps, so that direct stitching will cause a certain degree of redundancy. The path topology of DenseNet is shown in Fig 3 (b), whose input of each layer can be expressed as:

$$B_\ell = H_\ell\left([B_0, B_1, \ldots, B_{\ell-1}]\right), \tag{4}$$

where $[\cdot]$ denote concatenation, $[B_0, B_1, \ldots, B_{\ell-1}]$ refers to the concatenation of the outputs of $0^{th}, \ldots, \ell\text{-}1^{th}$ layer.

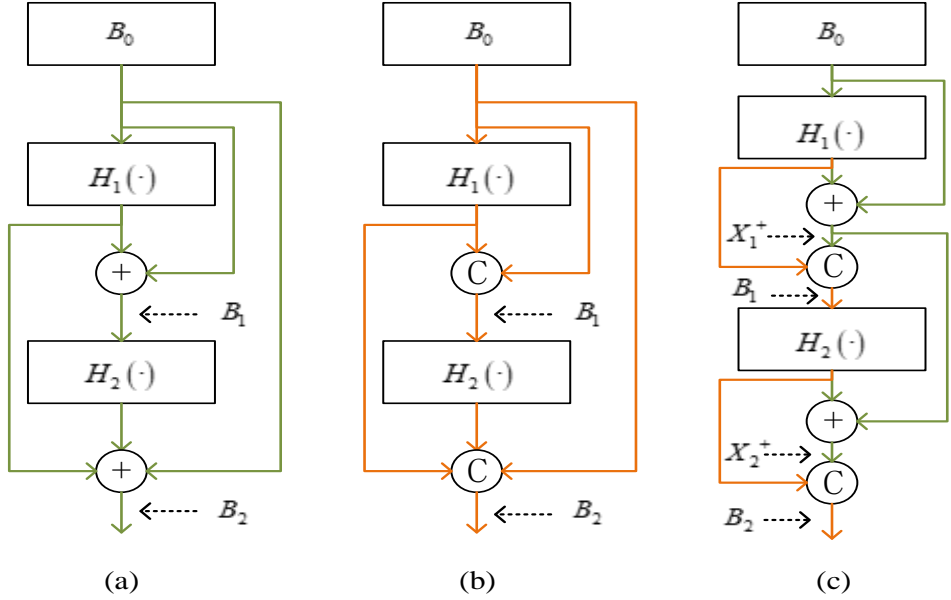

(a)      (b)      (c)

**Figure 3:** The topology of different networks, where the symbol "+" denote element-wise addition, and the letter "C" denote concatenation. (a) represents the topological structure of ResNet, using add between blocks. (b) shows the structure of DenseNet. (c) is HENet structure. Best in color.

HENet combined the advantages of ResNet and DenseNet, use element-wise addition and concatenation at the same time, create a new way of connection. Firstly, the outputs of the $\ell-1^{th}$ layer and the outputs of the $\ell^{th}$ layer are added, we denote as $X_\ell^+$. Then the outputs of the $\ell^{th}$ layer are concatenated to form the inputs of next layer. This way preserves the cross-layer connection to facilitate the gradient flow and performs different operations on the output of the same layer. Finally, them are put together to increase the utilization rate of the feature maps. Because it is not a direct stitching between layers, to some extent this can alleviate the problem of information redundancy. Its structural connection is shown in Fig 3 (c), in which $X_\ell^+$ can be expressed as:

$$X_\ell^+ = H_\ell(B_{\ell-1}) + X_{\ell-1}^+. \tag{5}$$

The outputs of HENet building block is:

$$B_{\ell} = \left[ X_{\ell}^{+}, H_{\ell}\left(B_{\ell-1}\right) \right]. \tag{6}$$

### 3.3 Network Architecture

Based on the above Stride 1 block and Stride 2 block in Fig 1 (a), we built a complete HENet, seen in Table 2. It is the Repeat 3 networks structure and will be used in the follow-up experiment. We choose to repeat the same number of times to build the network in the process of building HENet. In additional, we have taken the following improvements:

**Selection of the input image size**. Taking the two CIFAR datasets [16] as an example, we use the 31×31 input size. Our experiment shows that the use of odd image resolution (after each reduction of resolution the best is also odd, such as the CIFAR dataset using $31 \to 15 \to 7 \to 3$). Because padding can be 0 in the stride=2 convolution layer for odd size, it can be more effective to reduce the amount of calculation. For this slight change in our small network, the single core CPU can bring near 10% of the speed boost, which is very significant. We divided the network structure into 4 stages, of which 1, 2, 3 phases are composed of Stride1 block and Stride2 block. Stride 1 block repeats multiple times, and the Stride 2 block used for dimensional reduction.

Table 2: HENet architecture

| Layer   | Output size | Block S | Repeat | Group |     |
|---------|-------------|---------|--------|-------|-----|
|         |             | S       |        | m     | n   |
| Image   | 31×31×3     |         |        |       |     |
| Conv 1  | 31×31×24    |         | 1      |       |     |
| Stage 1 | 31×31×24    | 1       | 3      | 6     | 4   |
|         | 15×15×48    | 2       | 1      | 6     | 4   |
| Stage 2 | 15×15×48    | 1       | 3      | 8     | 6   |
|         | 7×7×96      | 2       | 1      | 8     | 6   |
| Stage 3 | 7×7×96      | 1       | 3      | 12    | 8   |
|         | 3×3×96      | 2       | 1      | 8     | 6   |
| Stage 4 | 1×1×192     | 2       | 1      | 12    | 8   |
| FC      |             |         | 10     |       |     |

**Cancel all pooling layers**. Starting from Network in Network [17], in many of the current neural network, the AVE pooling layer is adopted to replace the FC layer after convolutional layers. For ImageNet it is generally the 7×7 resolution, and then directly output to softmax layer. We think that pooling is a low efficiency of dimensional reduction method and is a waste of high dimensional features. So here we use a single Stride 2 block to do the last level of dimensional reduction and take this step as Stage 4. To ensure the number of final output channels, we do not double the number of channels in the Stride 2 block of Stage 3. Instead, it will double the channel in Stage 4. In addition, the resolution reducing in the whole network is implemented by Stride = 2, which avoids the computational redundancy caused by convolution calculation and pooling.

We can modify the depth of our network by changing the number of repeat at different resolutions, and then the groups settings are set using the rules mentioned in section 3.1, where the two parameters *m, n* depends on the number of input channels.

## 4. Experiments

As mentioned earlier, we have built some small network structures, and conducted experiments on small resolutions and small-size input datasets, among which the most famous datasets are CIFAR-10/100, SVHN [22]. Although further tuning can make our network more effective, in order to focus on the efficiency of the model itself, we used the same structure and the same training strategy in all the training datasets of the experiments, which will be explained later. This makes HENet easier to be used by other users without too much consideration for parameters optimization.

We have experimented with different repeat times of the model and compared with the same repeat number of ShuffleNet. In the 3rd section we have mentioned that, because our Stride 1 block input $k$ channels and output $2k$ channels, to ensure the comparison of result, we change ShuffleNet factor from 4 to 2. Channels increases as the sequence of {24, 48, 96, 192}. The following comparative experiment was done. The mean accuracy is the average accuracy of the last 5 times after the network structure convergence on the test set. The time data is that we randomly input a picture chosen from test dataset, and then execute forward propagation 1000 times on the single CPU core. The same model tests 5 times to obtain the final average run time.

### 4.1 Training

We used Caffe [18] framework for all experiments. We maintain the most part of training strategy, where we set the learning rate to 0.01, and use multistep learning rate transformation method. The maximum number of iterations is 65K, and learning rate is reduced with the multiple of 0.1 in the 32k, 48k respectively. Weight decay = 0.0005; mini-batch size = 128. Nesterov momentum [19] is used as an optimizer for parameter updates.

### 4.2 Datasets

**CIFAR-10:** CIFAR-10 is well known as a data set to get started in the field of computer vision, which is one of our key experiments. The CIFAR-10 datasets consist of 60000 32×32 colour images in 10 classes, with 6000 images per class, which are divided into 50000 training images and 10000 test images in our experiments. We use the data augmentation method in [6, 20, 21] in the two network training process, that is padding 4 pixels on each side and then a 31×31 randomly cropping, and mirror flip. A series of experiments are conducted. Experiment results are shown in Table 3.

Table 3: Comparison of results on CIFAR-10 dataset

| Name | Type | CIFAR-10 | | | |
|---|---|---|---|---|---|
| | | Mean accuracy | #param | #MFLOPS | Time(s) |
| HENet | Repeat 2 | 88.67 % | 507K | 7.3 | 3.524 |
| | Repeat 3 | 89.40 % | 641K | 10.2 | 4.846 |
| | Repeat 4 | 89.83 % | 775K | 13.2 | 6.184 |
| ShuffleNet | Repeat 2 | 87.07 % | 357K | 6.1 | 4.772 |
| | Repeat 3 | 88.03 % | 515K | 8.5 | 6.589 |
| | Repeat 4 | 88.23 % | 674K | 10.9 | 8.344 |
| | Repeat 5 | 88.57 % | 833K | 13.3 | 10.156 |

To increase the use of feature maps, the average pooling layer is not used in the final stage, so additional parameters were introduced, but the increase in the parameters on the network structure of this scale is acceptable. Under the same repeat structure, we have increased the speed by 26% than ShuffleNet in single CPU core, and the accuracy is higher.

From the view of same recognition accuracy, the result of ShuffleNet repeat 5 is equivalent to HENet repeat 2, but the actual running time of HENet is only 34.7% of ShuffleNet, the speed has been increased nearly twice times. At this time, the corresponding parameter quantity is 60.9%, and the theoretical calculation quantity is 54.9%. From these data, we can know that the actual running speed is higher than the theoretical running speed, which shows that the network structure of this paper is more advantageous to the actual inference operation on CPU.

**CIFAR-100:** This dataset is just like the CIFAR-10, except it has 100 classes containing 600 images each. There are 500 training images and 100 testing images per class. The 100 classes in the CIFAR-100 are grouped into 20 super classes. On this dataset we also use the same training strategy with CIFAR-10 to compare the performance of HENet with ShuffleNet in the larger class classification problem. Results are shown in Table 4.

Table 4: Comparison of results on CIFAR-100 dataset

| Name | Type | CIFAR-100 | | | |
|---|---|---|---|---|---|
| | | Mean accuracy | #param | #MFLOPS | Time(s) |
| HENet | Repeat 2 | 61.41 % | 524K | 7.3 | 3.53 |
| | Repeat 3 | 62.85 % | 658K | 10.2 | 4.85 |
| | Repeat 4 | 63.42 % | 792K | 13.2 | 6.85 |
| ShuffleNet | Repeat 2 | 60.53 % | 374K | 6.1 | 4.77 |
| | Repeat 3 | 61.30 % | 532K | 8.5 | 6.56 |
| | Repeat 4 | 62.89 % | 691K | 10.9 | 8.40 |
| | Repeat 5 | 63.29 % | 850K | 13.3 | 10.20 |

Similarly, in CIFAR-100, for the same recognition accuracy, ShuffleNet repeat 3 results is equal to HENet repeat 2, but HENet actual running time is only 53.8% of ShuffleNet, that is nearly twice speedup. The corresponding parameter quantity is 98.4%, and the theoretical calculation quantity is 71.6%. The results show that the advantage of HENet is reduced in the case of large class number. But our network structure is still conducive to the actual inference operation of CPU.

**SVHN:** SVHN is a real-world image dataset for developing machine learning and object recognition algorithms with minimal requirement on data preprocessing and formatting. We also use 31×31 size images as input, and do not use any form of data augmentation for this dataset. Experiment results are shown in Table 5. The overall performance of HENet is similar to the performance on CIFAR-100.

Table 5: Comparison of results on SVHN dataset

| Name | Type | SVHN | | | |
|---|---|---|---|---|---|
| | | Mean accuracy | #param | #MFLOPS | Time(s) |
| HENet | Repeat 2 | 93.86 % | 507K | 7.3 | 3.519 |
| | Repeat 3 | 94.71 % | 641K | 10.2 | 4.855 |
| | Repeat 4 | 95.03 % | 775K | 13.2 | 6.216 |
| ShuffleNet | Repeat 2 | 94.25 % | 357K | 6.1 | 4.761 |
| | Repeat 3 | 94.87 % | 515K | 8.5 | 6.555 |
| | Repeat 4 | 94.78 % | 674K | 10.9 | 8.450 |
| | Repeat 5 | 95.09 % | 833K | 13.3 | 10.25 |

## 5 Conclusions

In this paper, we firstly introduce the structural characteristics of some efficient networks and discuss the relationship between them. Then, we do some analysis and propose a novel network structure, called HENet, whose network structure uses a different number of groups, and is combined with channel shuffle operation. HENet can increase the utilization rate of feature map, and also effectively reduce the amount of calculation. Finally, the experimental results show that our model's efficiency is more than 1 times higher than ShuffleNet on many open source datasets.

## 6 Future Work

In this paper, we only discuss the experimental results on small datasets. In the future, we shall construct a larger network structure and conduct more experiments on the large resolution datasets, such as ImageNet, including the different resolution of the number of repeat, and trying other tasks, such as object detection, instance segmentation and other fields of application.